\documentclass[runningheads]{llncs}
\usepackage[T1]{fontenc}
\usepackage{graphicx,verbatim}
\usepackage{amsmath}
\usepackage{bm}
\usepackage{layouts}
\usepackage{hyperref}
\usepackage{array}
\usepackage{booktabs}
\usepackage{multirow}
\usepackage{multicol}
\usepackage{makecell}
\usepackage{xcolor}

\begin{document}
\newcommand{\pd}{\mathrm{PD}}
\newcommand{\Tone}{\mathrm{T}_1}
\newcommand{\Ttwo}{\mathrm{T}_2}
\newcommand{\phys}{\mathbf{z}}
\newcommand{\inv}{\mathrm{INV}}
\newcommand{\tinv}{t_{\mathrm{INV}}}
\newcommand{\TR}{\mathrm{TR}}
\newcommand{\TE}{\mathrm{TE}}
\newcommand{\TD}{\mathrm{TD}}
\newcommand{\TS}{\mathrm{TS}}
\newcommand{\Eone}{\mathrm{E}_1}
\newcommand{\Etwo}{\mathrm{E}_2}
\newcommand{\sig}[1]{f_{\mathrm{#1}}}
\newcommand{\mulgauss}{\mathcal{N}(0, \mathbf{I})}
\title{Reverse Imaging for Wide-spectrum Generalization of Cardiac MRI Segmentation}
\titlerunning{Reverse Imaging for CMR Segmentation}

\author{Yidong Zhao\inst{1}, Peter Kellman\inst{2}, Hui Xue\inst{3}, Tongyun Yang\inst{1}, Yi Zhang\inst{1}, \\Yuchi Han\inst{4}, Orlando Simonetti\inst{4}, Qian Tao\inst{1}}
\authorrunning{Y. Zhao et al.}
\institute{Department of Imaging Physics, Delft University of Technology, Delft, The Netherlands \and
National Heart Lung and Blood Institute, National Institutes of Health, Bethesda, Maryland, USA \and
Microsoft Research \and
Cardiovascular Division, The Ohio State University Wexner Medical Center,
Columbus, Ohio, USA\\
    \email{q.tao@tudelft.nl}}

\maketitle              
\begin{abstract}
Pretrained segmentation models for cardiac magnetic resonance imaging (MRI) struggle to generalize across different imaging sequences due to significant variations in image contrast. These variations arise from changes in imaging protocols, yet the same fundamental spin properties, including proton density, $\Tone$, and $\Ttwo$ values, govern all acquired images. With this core principle, we introduce \emph{Reverse Imaging}, a novel physics-driven method for cardiac MRI data augmentation and domain adaptation to fundamentally solve the generalization problem. Our method reversely infers the underlying spin properties from observed cardiac MRI images, by solving ill-posed nonlinear inverse problems regularized by the prior distribution of spin properties. We acquire this ``spin prior'' by learning a generative diffusion model from the multiparametric SAturation-recovery single-SHot acquisition sequence (mSASHA) dataset, which offers joint cardiac $\Tone$ and $\Ttwo$ maps. Our method enables approximate but meaningful spin-property estimates from MR images, which provide an interpretable ``latent variable'' that lead to highly flexible image synthesis of arbitrary novel sequences. We show that Reverse Imaging enables highly accurate segmentation across vastly different image contrasts and imaging protocols, realizing wide-spectrum generalization of cardiac MRI segmentation.


\end{abstract}
\section{Introduction}
Segmentation is crucial for evaluating cardiac biomarkers from cardiac magnetic resonance imaging (MRI), for example, the ejection fraction from balanced steady-state free-precession (bSSFP) cine imaging~\cite{bernard2018deep,campello2021multi}. Learning-based methods have become the standard practice for cardiac MRI segmentation~\cite{isensee2021nnu,bernard2018deep,campello2021multi,martin2023deep}, mostly trained on widely accessible bSSFP cine images with manual annotations~\cite{bernard2018deep,campello2021multi,martin2023deep}. However, models trained on bSSFP cine are prone to fail in test images with heterogeneous contrasts~\cite{campello2021multi,martin2023deep}. This can be partially mitigated by data augmentation (e.g. blur, gamma correction, etc.)~\cite{isensee2021nnu}, which effectively generalizes pretrained models to bSSFP cine images cross vendors and centers~\cite{tao2019deep,campello2021multi,chen2022enhancing}. 

However, a much stronger contrast change comes from different MRI sequences, which cannot be tackled by common image augmentation techniques. For example, an alternative to bSSFP is the gradient echo sequence (GRE), which is less affected by field inhomogeneities~\cite{ridgway2010cardiovascular} and more suitable for patients with implanted devices~\cite{ferreira2013cardiovascular}. However, GRE has much poorer contrasts and easily fails the models trained on bSSFP. An even stronger contrast change occurs in quantitative imaging, such as $\Tone$ mapping by the modified Look-Locker Inversion Recovery Sequence (MOLLI)~\cite{messroghli2004modified}. In MOLLI, the image readouts have substantial contrast variations during the spin relaxation process, alternating between bright and dark blood, with low myocardium-blood contrast. Consequently, segmentation models trained on bSSFP images (source domain) can fail completely on MOLLI images (target domain) even with extensive data augmentation.

To tackle the cross-sequence generalization challenge, previous works have focused on disentangling ``\textit{content}'' and ``\textit{style}''~\cite{ouyang2019data,ouyang2022causality}. BayeSeg learns to remove the style and extracts the boundary for segmentation~\cite{gao2022joint,gao2023bayeseg}. Domain adaptation techniques strive to learn content and style embeddings~\cite{chen2020unsupervised,xie2022unsupervised}. The learned ``content'' is supposed to be shared between sequences, while cross-sequence translation is achieved by interchanging the learned ``style'' \cite{chen2020unsupervised,xie2022unsupervised,cui2021structure}. However, such strategies require data from new sequences in order to learn the disentangling, and, importantly, the extracted content or style is barely interpretable. In this study, we argue that the fundamental but overlooked ``content'' is the underlying \emph{spin properties}, including the voxel-wise proton density (PD), $\Tone$, and $\Ttwo$~\cite{hashemi2012mri} of blood and tissue. Given this content, the style is fully governed by the MRI signal models of new pulse sequences. This motivates us to infer the spin properties from the observed image, for more interpretable and physics-grounded image translation to tackle the cross-sequence domain generalization challenge.  

Unlike quantitative MRI that directly maps spin properties~\cite{messroghli2004modified,kellman2014t1,chow2022improved}, retrieving the underlying spin properties from a qualitative image such as bSSFP is ill-posed because infinitely many solutions may explain the same image. This challenge parallels that of accelerated MRI, where partial measurements are insufficient for image reconstruction, yet the inverse problem can be solved by incorporating a regularizer that defines the prior distribution of plausible images~\cite{pauly2008compressed}. Recent MRI reconstruction works use generative diffusion models~\cite{ho2020denoising} as \emph{a priori} regularizers, allowing high-quality reconstruction from partial measurements~\cite{song2021solving,chung2022diffusion}. Following the same spirit, we introduce a diffusion-based generative prior on spin properties, called ``spin prior'', to regularize the ill-posed inverse problem of spin property estimation. Specifically, we leverage the multiparametric SAturation-recovery single-SHot acquisition (mSASHA) datasets, which perform joint cardiac $\Tone$ and $\Ttwo$ mapping, to learn the spin prior. We term the process of inferring spin properties from observed images as \textit{Reverse Imaging}. Reverse Imaging uniquely enables physics-grounded data augmentation and domain adaptation without the need for target domain data. We will show that Reverse Imaging leads to high-quality zero-shot generalization to a wide spectrum of MRI sequences. We make the following contributions:
\begin{itemize}
    \item We propose an interpretable cross-sequence translation framework for cardiac MRI, which explicitly formulates the translation as applying MRI physics forward models to estimated tissue spin properties (i.e. $\pd$, $\Tone$ and $\Ttwo$).
    \item We introduce Reverse Imaging, a novel approach that reversely infers spin properties from observed MR images using a physics-guided reverse diffusion process. The diffusion model serves as a generative prior that regularizes the estimation of the underlying spin properties.
    \item With Reverse Imaging, we achieve high-quality zero-shot generalization of cardiac MRI segmentation, from bSSFP to a wide spectrum of unseen contrasts and imaging protocols.
\end{itemize}

\section{Method}

\begin{figure}[t]
    \centering
    \includegraphics[width=\linewidth]{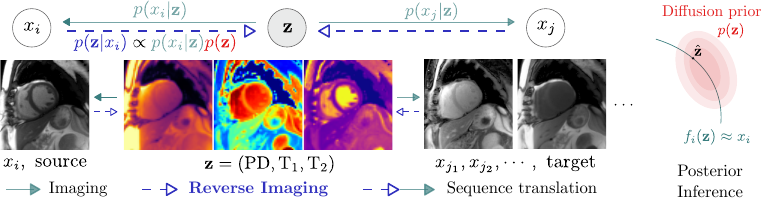}
    \caption{Images acquired by sequences $i$ and $j$ from a subject share the same underlying spin properties $\phys$. Reverse Imaging (dashed arrows) estimates $\hat{\phys}$ from the prior $p(\phys)$ that explains the observed image $x_i$. Images of an arbitrary sequence can be generated by MR physics from $\hat{\phys}$ (solid arrows). In this process, the only unknown is the prior $p(\phys)$, which we learn from the mSASHA data by a diffusion model (red).}
    \label{fig:method-overview}
\end{figure}
\subsection{Physics-based Cross-Sequence Translation}
\subsubsection{Physics-Prior Decomposition} Let $x_i$ and $x_j$ be images acquired using sequences $i$ and $j$, on which we aim to perform cross-sequence translation. The direct translation $p(x_j|x_i)$ from $x_i$ to $x_j$ is intractable. To solve this, we propose a decomposition of $p(x_j|x_i)$, by introducing an underlying physics $\phys=(\pd, \Tone, \Ttwo)$ as a ``latent variable'' :
\begin{align}
    p(x_j|x_i)&=\int_\phys p(x_j, \phys|x_i)d\phys =\int_\phys p(x_j|\phys, x_i)p(\phys|x_i) d\phys = \int_\phys p(x_j|\phys) p(\phys|x_i)d\phys. \label{eq:bayes_decomp}
\end{align}
A probability graph model is shown in Fig.~\ref{fig:method-overview}. Reverse Imaging can be formulated as posterior inference $p(\phys|x_i)$, expressed as:
\begin{align}
    \log p(\phys|x_i) \propto \log p(x_i|\phys) + \log p(\phys) + \mathrm{Const} .
    \label{eq:log-post}
\end{align}
As a result of~\eqref{eq:bayes_decomp} and~\eqref{eq:log-post}, the inference $p(x_i|x_j)$ relies on two likelihoods $p(x_i|\phys)$ and $p(x_j|\phys)$, which are directly accessible from the imaging physics forward model. The only missing piece for inference, therefore, is the prior distribution on the spin properties $p(\phys)$. Moreover, we can interchange $x_i$ and $x_j$ in \eqref{eq:bayes_decomp} once the prior $p(\phys)$ is known, allowing dual-direction translation of both $p(x_j|x_i)$ and $p(x_i|x_j)$. 

\subsubsection{Likelihoods from MR Imaging Physics} The log-likelihood of observing image $x_i$ from spin properties $\phys$ under sequence $i$ is: 
\begin{align}
    \log p(x_i|\phys) \propto \left \Vert f_i \left(\phys\right) - x_i \right\Vert_2^2,
\end{align}
where the signal equation (i.e. forward model) $f_i$ can be expressed by the Bloch equation of MR imaging~\cite{hashemi2012mri}. Below, we list the signal equations for the sequences involved in this work, namely, bSSFP, MOLLI $\Tone$ mapping, and GRE. We denote the flip angle (FA) as $\omega$, the repetition and echo times as $\TR$ and $\TE$. The bSSFP imaging equation ~\cite{scheffler2003principles} with short $\TR$ and $\TE$ is: 
\begin{align}
    \sig{SS} (\phys) = \frac{\pd \sin(\omega)}{1+\cos(\omega)+[1-\cos(\omega)]\Tone/\Ttwo}. \label{eq:imaging-ssfp}
\end{align}
For $\Tone$ mapping, the phase-insensitive readout~\cite{schmitt2004inversion} at inversion time $\tinv$ is:
\begin{align}
    \sig{MOLLI}(\phys) &= \left\vert\sig{SS}(\phys) \left(1- \inv \cdot \exp({-\frac{\tinv}{\Tone^*}})\right) \right\vert, \label{eq:imaging-molli}
\end{align}
where the apparent $\Tone$ is $\Tone^* = \left(\frac{1}{\Tone}\cos^2(\frac{\omega}{2}) + \frac{1}{\Ttwo}\sin^2(\frac{\omega}{2})\right)^{-1}$ and the inversion factor $\inv =1+\frac{\sin(\frac{\omega}{2})}{\sin(\omega)}\left(1+\cos(\omega)+[1-\cos(\omega)]\frac{\Tone}{\Ttwo}\right)$. For GRE cine images from patients with implanted devices, let $\Eone=\exp(-\frac{\TR}{\Tone})$ and $\Etwo=\exp(-\frac{\TE}{\Ttwo})$, the GRE signal model~\cite{hashemi2012mri} is described as 
\begin{align}
    \sig{GRE}(\phys) &= \pd\sin(\omega)\frac{1-\Eone}{1-\cos(\omega) \Eone} \Etwo.\label{eq:imaging-gre}
\end{align}

\subsection{Generative Prior on Spin Properties from mSASHA}
To learn the spin prior $p(\phys)$, we leverage an advanced quantitative MRI sequence mSASHA, which enables joint $\Tone$ and $\Ttwo$ mapping~\cite{chow2022improved}. mSASHA consists of a train of saturation recovery pulses (SR) with SR time $\TS$, followed by a $\Ttwo$-prep pulse with duration $\TD$ and echo time $\TE$. Its signal equation is: 
\begin{align}
    \sig{mSASHA}(\phys) = A(\phys)\left\{1 - \left[1 - \left(1 - e^{-\frac{\TS}{\TE}} \right)e^{-\frac{\TE}{\Ttwo}} \right]e^{-\frac{\TD}{\Tone}} \right\},
\end{align}
where the unknowns $A(\phys), \Tone, \Ttwo$ can be extracted from multiple acquisitions of $\sig{mSASHA}$ with varying $\TS$, $\TD$ and $\TE$ by least-square fitting~\cite{chow2022improved}. $A$ encodes $\pd$ and sequence parameters, and the approximation $\pd\approx A$ can be used~\cite{akccakaya2015improved,akccakaya2016joint}. 

To model the prior of spin properties $\phys=(\pd,\Tone,\Ttwo)$, we propose to train a denoising diffusion probabilistic model (DDPM)~\cite{ho2020denoising} in the $\phys$ space. DDPM adds noise to the real $\phys$'s from mSASHA in the forward process $\phys_t=\sqrt{1-\beta_t}\phys_{t-1}+\sqrt{\beta_t}\mathbf{\epsilon}$ for $T$ steps, where $\beta_t$ is the noise scheduler at $t$ and $\mathbf{\epsilon}\sim\mulgauss$. The learned model can generate data by starting with a Gaussian distribution $\phys_T\sim\mulgauss$ and reversing the diffusion process:
\begin{align}
    \phys_{t-1} &= \frac{1}{\sqrt{\alpha_t}}\left(\phys_t + (1-\alpha_t)\nabla_{\phys_t} \log p (\phys_t)\right) +\sigma_t\mathbf{\epsilon} ,
    \label{eq:reverse_sde}
\end{align}
where $\alpha_t=1-\beta_t$ and $\sigma_t^2=\frac{1-\overline{\alpha}_{t-1}}{1-\overline{\alpha}_t}\beta_t$. The diffusion model learns a denoising backbone parameterized by $\theta$, $\nabla_{\phys_t} \log p (\phys_t)\approx -\frac{1}{\sqrt{1-\overline{\alpha}_t}}\epsilon_\theta(\phys_t, t)$, $\overline{\alpha}_t=\prod_{r=1}^{t} \alpha_r$, as the generative prior for $\phys$. 

\subsection{Reverse Imaging and Cross-Sequence Synthesis}
Reverse Imaging searches for $\phys$ in the prior space that best explains the observation $x_i$. Following conditional generation principles~\cite{chung2022diffusion}, we replace $\nabla_{\phys_t} \log p (\phys_t)$ in~\eqref{eq:reverse_sde} with $\nabla_{\phys_t} \log p(\phys_t|x_i)$:
\begin{align}
    \nabla_{\phys_t} \log p(\phys_t|x_i) \propto \nabla_{\phys_t}  \log p(x_i|\phys_t) + \nabla_{\phys_t}  \log p (\phys_t ).
\end{align}
We approximate the time-step-dependent likelihood $\log p(x_i|\phys_t)$ by: $\log p(x_i|\phys_t)\approx\log p(x_i|\tilde{\phys}_0(\phys_t))$, where $\tilde{\phys}_0(\phys_t) \approx (\phys_t-\sqrt{1-\overline{\alpha}_t} \epsilon_\theta(\phys_t))/\sqrt{\overline{\alpha}_t}$~\cite{chung2022diffusion}.
Starting from $\phys_T\sim\mulgauss$, the recursive process for reverse the imaging sequence $x_i$ is 
\begin{align}
    \phys_{t-1} &= \frac{1}{\sqrt{\alpha_t}}\left(\phys_t - \frac{1-\alpha_t}{\sqrt{1-\overline{\alpha}_t}}\epsilon_\theta(\phys_t, t)\right) +\sigma_t\mathbf{\epsilon}  -\xi \frac{\partial}{\partial \phys_t} \left \Vert f_i \left(\tilde{\phys}_0(\phys_t)\right) - x_i \right\Vert_2^2,
    \label{eq:dps-inverse}
\end{align}
where $\xi$ is a step size parameter. The guided reverse diffusion process in~\eqref{eq:dps-inverse} yields the estimated spin properties $\hat{\phys}=\phys_0$ in the prior distribution that reconstructs $x_i$. After Reverse Imaging, we can synthesize images of novel sequences using the imaging physics equations $f_j$ described in~\eqref{eq:imaging-ssfp}-\eqref{eq:imaging-gre}. Synthesizing other imaging sequences is also possible by sequence-specific Bloch equations. In addition, we can also use the image physics $f_j$ of $x_j$ to guide generation by replacing $f_i$, $x_i$ with $f_j$, $x_j$ in~\eqref{eq:dps-inverse} and translating from target to source (T2S).

\section{Experiments}
\subsubsection{Datasets} We used the ACDC dataset~\cite{bernard2018deep} training split with bSSFP cine images of 100 subjects as the source domain dataset. The mSASHA dataset for learning the spin prior consists of 63 subjects with 186 short-axis slices. For evaluation of cross-sequence generalizability, we included two datasets imaged with different sequences: 1) MOLLI $\Tone$ mapping data: 49 subjects with 131 short-axis slices in total. Each slice contains 11 baseline images with different contrasts. 2) cine MRI of 25 subjects with implanted devices: 16 subjects with the bSSFP sequence (Dev-bSSPF) and 9 with GRE sequence (Dev-GRE). Both the MOLLI and the Device datasets have significantly different contrasts than those of the ACDC.

\subsubsection{Experimental Settings}
The DDPM is pretrained on ACDC images for $36,000$ steps and then fine-tuned with the real spin properties of from mSASHA for $6,000$ steps on the resolution of $128\times128$. We employ the hugging-face \texttt{diffuser} for the implementation of DDPM~\cite{von-platen-etal-2022-diffusers}. The estimated spin properties were resized to their original resolution after Reverse Imaging. We perform $T=1000$ forward steps and $200$ reverse steps of diffusion in~\eqref{eq:dps-inverse} for Reverse Imaging, with the step size $\xi=400$. As the exact FA for ACDC scans is unknown, we approximate it with $\omega=45^{\circ}$ as typically used. We open source the estimated spin properties for the ACDC dataset and code\footnote[1]{\url{https://github.com/Ido-zh/cmr_reverse.git}}. 
For evaluating the generalization performance, we include the following methods:
\begin{itemize}
    \item \textbf{Baseline} We train an nnUNet\cite{isensee2021nnu} with the ACDC data as the baseline;
    \item \textbf{BayeSeg} We compare our model against BayeSeg~\cite{gao2022joint,gao2023bayeseg} which extracts the boundary and contour information from images for generalization; 
    \item  \textbf{RI-T2S} We translate the MOLLI and GRE data into bSSFP cine by Reverse Imaging (RI) and perform segmentation on the translated (T2S) images with the baseline model;
    \item \textbf{RI-Aug.} We perform Reverse Imaging on the ACDC cine images and extend the nnUNet framework by synthesizing MOLLI and GRE images through the estimated spin properties for physics-based augmentation in~\eqref{eq:imaging-molli} and~\eqref{eq:imaging-gre}.
\end{itemize}

\section{Results and Discussion}
\subsection{Reverse Imaging}
\begin{figure}[htb!]
    \centering
    \includegraphics[width=0.95\linewidth]{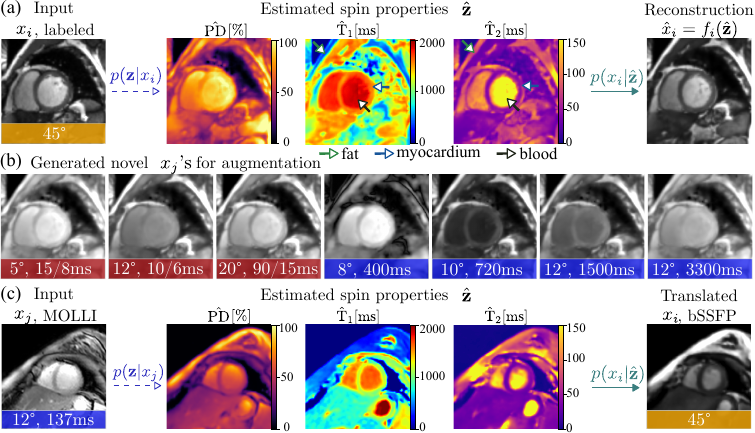}
    \caption{(a) Reverse Imaging a bSSFP image $x_i$ (yellow). The estimated $\hat{\phys}$ accurately reconstructs $\hat{x}_i\approx x_i$. (b) Generating GRE (red) and MOLLI (blue) images with $\hat{\phys}$. The FA in degrees and $\TR/\TE$ or $\tinv$ in milliseconds are given for each image. (c) Reverse Imaging a MOLLI readout for target-to-source (T2S) translation. }
    \label{fig:vis-ssfp-inv-gen}
\end{figure}

We first show an example of Reverse Imaging a bSSFP cine in Fig.~\ref{fig:vis-ssfp-inv-gen}-(a). The estimated spin properties $\hat{\phys}$ accurately reconstruct the bSSFP image, indicating a high likelihood. The predicted PD has low contrast between tissues, while in $\Tone$ and $\Ttwo$ estimation, blood has a high $\Tone$ and $\Ttwo$, myocardium has high $\Tone$ but low $\Ttwo$, and fat has low $\Tone$ but a high $\Ttwo$. These match the reference relative magnitude of the spin properties~\cite{xu2023reference}. Based on estimated spin properties, images of various imaging protocols can be generated with large contrast variations, as shown in Fig.~\ref{fig:vis-ssfp-inv-gen}-(b). Reverse Imaging does not provide precise spin property estimation, but we note that they are not used for quantitative evaluation purposes w.r.t. $\Tone$ or $\Ttwo$, instead, they serve to synthesize novel contrasts that are essential for wide-spectrum generalization. Reverse Imaging also enables the T2S translation from MOLLI to bSSFP cine (Fig.~\ref{fig:vis-ssfp-inv-gen}-(c)). 

\subsection{Segmentation Performance}
\label{sec:res-seg-performance}
\begin{figure}[htb!]
    \centering
    \includegraphics[width=\linewidth]{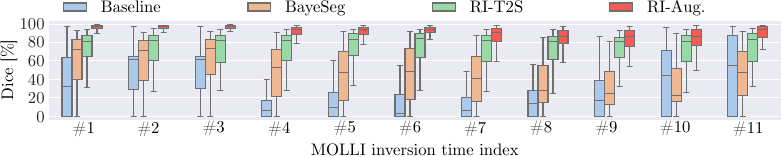}
    \caption{Segmentation accuracy on each MOLLI readout during inversion recovery.}
    \label{fig:molli-seg-dice}
\end{figure}

\begin{figure}[htb!]
    \centering
    \includegraphics[width=\linewidth]{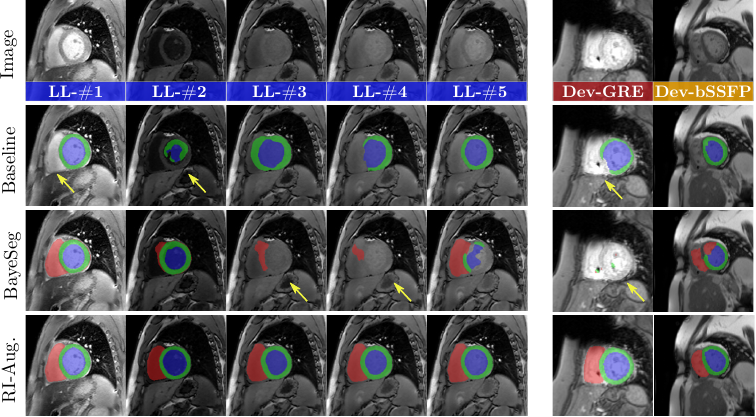}
    \caption{Segmentation of 5 MOLLI baseline images (LL-\#1-\#5) and 2 device cine images, all of which have different contrast than bSSFP cine images used for training.}
    \label{fig:seg-qualitative}
\end{figure}

We now demonstrate that Reverse Imaging can significantly improve the generalization of segmentation models. In Table~\ref{table:dice_results}, we list the segmentation accuracy measured by the Dice score of the left ventricle (LV), myocardium (MYO) and right ventricle (RV) in the challenging MOLLI and Device datasets. To further evaluate the robustness against contrast variation, we also report the Dice for each baseline image of MOLLI in Fig.~\ref{fig:molli-seg-dice}, as MOLLI readouts show strong contrast variation during inversion recovery. Qualitative examples of segmentation results are shown in Fig.~\ref{fig:seg-qualitative}.

\begin{table}[htb!]
\caption{Segmentation accuracy on novel sequences measured by Dice score [$\%$]. Statistical significance ($p<0.05$) is indicated by $^*$.}
\label{table:dice_results}
\centering
\begin{tabular}{p{1.6cm}<{\centering}p{1.7cm}<{\centering}p{1.7cm}<{\centering}p{1.7cm}<{\centering}p{1.7cm}<{\centering}p{1.7cm}<{\centering}p{1.5cm}<{\centering}  }
\toprule
\multirow{2}{*}{\textbf{Methods}} & \multicolumn{3}{c}{\textbf{MOLLI}} & \multicolumn{3}{c}{\textbf{Device}} \\  
\cmidrule(lr){2-4}\cmidrule(lr){5-7}
                         & RV & MYO & LV  & RV & MYO & LV \\
\midrule
Baseline~\cite{isensee2021nnu} & $24.0\pm12.9$ & $47.4\pm19.5$ & $39.9\pm22.0$ & $68.9\pm25.2$ & $86.9\pm8.6$ & $91.9\pm7.6$ \\
BayeSeg~\cite{gao2023bayeseg} & $51.5\pm26.9$ & $41.4\pm19.9$ & $57.0\pm24.1$ & $53.6\pm21.0$ & $62.6\pm12.6$ & $72.6\pm19.1$ \\
RI-T2S & $63.6\pm36.5$ & $69.8\pm20.1$ & $82.7\pm21.0$ & $66.4\pm24.2$ & $82.3\pm15.2$ & $88.6\pm12.1$ \\
RI-Aug. & $\mathbf{87.0\pm17.8}^*$ & $\mathbf{86.5\pm9.8}^*$ & $\mathbf{91.6\pm9.3}^*$ & $\mathbf{87.4\pm9.5}^*$ & $\mathbf{93.1\pm4.0}^*$ & $\mathbf{96.0\pm 3.5}^*$ \\

\bottomrule
\end{tabular}
\end{table}
 

The baseline nnUNet generalizes poorly in the MOLLI images (Table~\ref{table:dice_results}), especially when the blood-myocardium contrast is inverted (see \#4-\#7 in Fig.~\ref{fig:molli-seg-dice} and LL-\#2 in Fig.~\ref{fig:seg-qualitative}). BayeSeg can improve performance on dark blood images, yet it can still fail on low-contrast images (e.g. Fig.~\ref{fig:seg-qualitative} LL-\#3-\#5). The device images have bright blood but lower contrast, and the baseline nnUNet works more robustly than BayeSeg with higher Dice scores (Table~\ref{table:dice_results}). Performing the T2S transform (RI-T2S) significantly improved the segmentation accuracy for MOLLI, but a performance drop is observed in the device dataset. This is because the short $\TE$ ($\approx1.5$ ms) in GRE minimizes $\Ttwo$ weighting (cf. \eqref{eq:imaging-gre}). In this case, Reverse Imaging is limited by the little amount of information carried by the observed image. However, this can be solved by using Reverse Imaging for augmentation instead (RI-Aug.), which achieves the highest segmentation accuracy in both datasets, significantly better than all other methods (Table~\ref{table:dice_results}, $p<0.05$). With RI-Aug., the trained model consistently achieves highly accurate segmentation on images with various contrasts, including low-contrast, bright and dark blood, and GRE images (Fig.~\ref{fig:seg-qualitative}). 

\section{Conclusion}

We introduced Reverse Imaging, a novel physics-grounded approach for cross-sequence generalization in cardiac MRI segmentation. Combining imaging physics and a generative diffusion prior, our method estimates underlying spin properties from observed images. Reverse Imaging provides an interpretable and physically grounded solution for domain adaptation, addressing the challenge of contrast variations across imaging protocols. Our experiments demonstrate that Reverse Imaging significantly boosts segmentation robustness, enabling zero-shot generalization to unseen sequences without requiring target-domain data. This approach offers a new paradigm for improving cardiac MRI segmentation generalizability, with potential applications to broader medical imaging tasks where domain shifts pose a significant challenge.

\begin{credits}
\subsubsection{\ackname}
We acknowledge the financial support from the TU Delft AI Initiative and the Dutch Research Council (NWO). \textcolor{blue}{This preprint has not undergone post-submission improvements or corrections.}

\subsubsection{\discintname}
We the authors hereby declare no competing interests in this paper. 
\end{credits}
%
%
%
\bibliographystyle{splncs04}
\bibliography{bibliography}
\end{document}